\documentclass[11pt]{article}

\usepackage[preprint]{acl}

\usepackage{times}
\usepackage{latexsym}

\usepackage[T1]{fontenc}
\usepackage{booktabs}
\usepackage{graphicx}
\usepackage{booktabs}
\usepackage{multirow}
\usepackage{graphicx}
\usepackage[table]{xcolor}
\usepackage{multirow}

\usepackage[utf8]{inputenc}

\usepackage{microtype}

\usepackage{inconsolata}

\usepackage{graphicx}

\title{SAGE: From Direct Answering to Evidence-Grounded Inference for Chinese Ancient Document Understanding}

\author{
Yuchuan Wu\textsuperscript{1} \quad
Xuan Luo\textsuperscript{1} \quad
Yinglian Zhu\textsuperscript{1} \quad
Meng Fang\textsuperscript{2} \quad
Xiangyang Xue\textsuperscript{1} \quad
Bin Li\textsuperscript{1}\thanks{Corresponding author.}
\\[2mm]
\textsuperscript{1}Fudan University
\qquad
\textsuperscript{2}University of Liverpool
\\[1mm]
\texttt{\{ycwu24, luox25, ylzhu22\}@m.fudan.edu.cn}
\\
\texttt{Meng.Fang@liverpool.ac.uk}
\\
\texttt{\{xyxue, libin\}@fudan.edu.cn}
}
\begin{document}
\maketitle

\begin{abstract}
Chinese ancient document understanding demands complex visual, linguistic, and historical reasoning. Current Large Vision-Language Models (LVLMs) typically rely on an opaque, single-pass generation paradigm, often producing overconfident and weakly grounded responses. To address this, we propose \textbf{SAGE}, an evidence-grounded multi-agent framework that reformulates Chinese ancient document understanding as evidence-grounded inference rather than direct answer generation. SAGE coordinates specialized agents for task-aware planning, tool-mediated evidence acquisition, claim-level verification, and bounded replanning under a constrained shared-state runtime. This design supports bounded evidence seeking, answer revision, and abstention when grounding is insufficient. Experiments on the AncientDoc benchmark show that SAGE consistently outperforms matched direct-answering baselines across three LVLM backbones. Remarkably, SAGE with Qwen3.5-9B surpasses much larger monolithic LVLMs on most evaluated metrics, highlighting the importance of structured, evidence-grounded inference beyond model scaling.
\end{abstract}

\section{Introduction}

\begin{figure*}[t]
    \centering
    \includegraphics[width=0.95\textwidth]{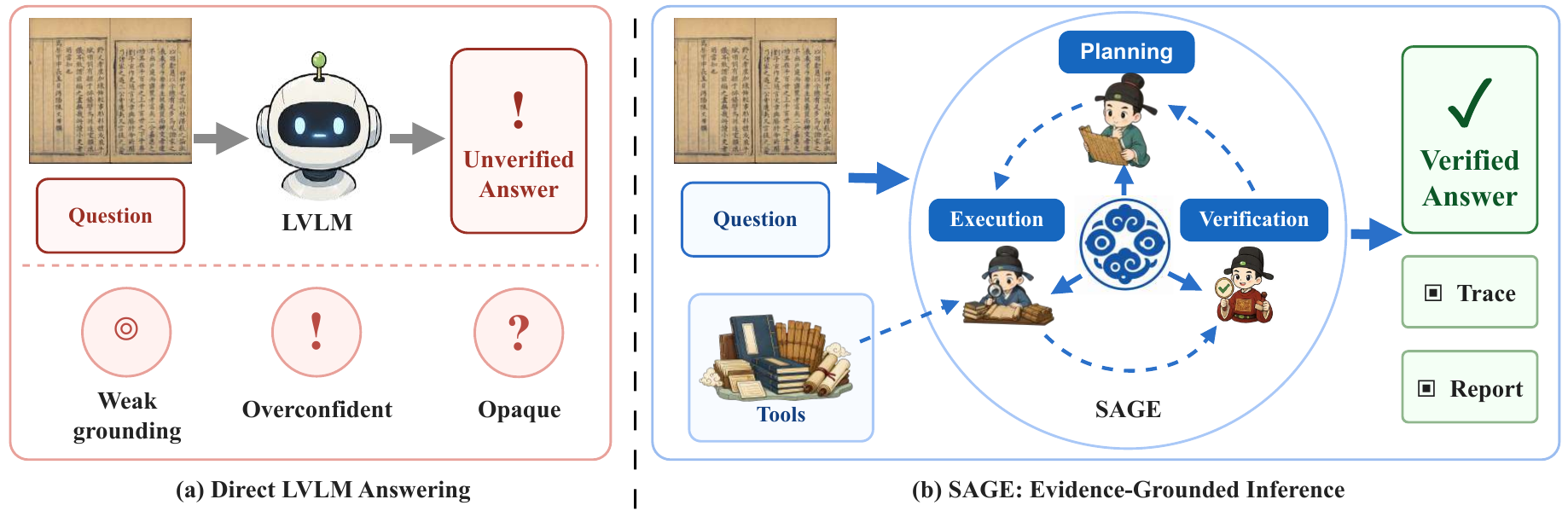}
    \caption{
Direct LVLM answering vs. SAGE: Evidence-Grounded Inference. 
}
    \label{fig:intro}
\end{figure*}

Chinese ancient documents preserve millennia of historical, literary, philosophical, and cultural knowledge~\citep{wei-etal-2024-ac,zhao2025fuxibenchmarkevaluatinglanguage,cao-etal-2024-tonggu}. While extracting information from these page images is crucial for the digital humanities and cultural heritage preservation, the task introduces complexities far beyond modern document understanding~\citep{mathew2020docvqa,NEURIPS2024_ae0e4328}. These texts feature vertical layouts, traditional and variant characters, missing punctuation, dense annotations, and highly implicit historical-cultural semantics. As highlighted by recent benchmarks like AncientDoc~\cite{yu2025benchmarkingvisionlanguagemodelschinese}, mastering this domain is inherently a multi-stage endeavor that requires models to seamlessly integrate visual reading, classical Chinese interpretation, and knowledge-grounded reasoning.

These compounding challenges severely restrict the ability of large vision-language models (LVLMs) to move beyond basic visual recognition toward faithful interpretation. Despite this, most existing LVLM-based systems still rely on a direct-answering paradigm~\citep{fu2024ocrbench,NEURIPS2024_ae0e4328}: given an image and a query, the model generates a response in a single forward pass. This approach collapses reading, evidence acquisition, reasoning, and verification into one opaque generation step. Consequently, the resulting answers are frequently overconfident, weakly grounded, and notoriously difficult to diagnose when errors occur.

In authentic scholarly practice, however, even human experts are not omniscient. Historians answer complex questions by systematically reading the text, identifying visual clues, consulting external references, and verifying claims—crucially expressing uncertainty when evidence is lacking. This reality motivates a necessary paradigm shift in ancient document understanding: moving away from monolithic, direct LVLM generation toward an evidence-grounded, verification-aware, and abstention-capable inference process.

To this end, we propose \textbf{SAGE}, an evidence-grounded multi-agent framework for Chinese ancient document understanding. Moving away from the conventional single-step generation paradigm, \textbf{SAGE} decomposes evidence-grounded inference into three role-specialized agents: (1) a \emph{Scholarly Planning Agent} that analyzes the query and formulates a task- and budget-aware evidence-seeking strategy; (2) a \emph{Tool-Augmented Execution Agent} that invokes constrained tools---such as page reading, text normalization, term extraction, and local evidence retrieval---to gather evidence and synthesize a grounded response; and (3) a \emph{Scholar Verifier Agent} that evaluates whether the generated claims are sufficiently supported by the acquired evidence. These agents operate under a constrained evidence-grounded runtime, which maintains a shared evidence state, regulates tool access and budget, records execution traces, and applies predefined control rules to stop, replan for additional evidence, or abstain when grounding is insufficient. Figure~\ref{fig:intro} illustrates the conceptual contrast between this evidence-grounded inference process and direct LVLM answering.

Extensive experiments on the AncientDoc benchmark across three LVLM backbones demonstrate that SAGE consistently surpasses matched direct-answering baselines on all evaluated metrics, yielding the most substantial gains on knowledge-intensive and linguistic tasks. Remarkably, SAGE equipped with the Qwen3.5-9B model achieves state-of-the-art performance on seven of the eight metrics, outperforming monolithic LVLMs that are orders of magnitude larger. This underscores a critical insight: evidence-grounded inference and rigorous verification can elevate ancient document understanding far more effectively than merely scaling base model parameters. Furthermore, trace-based analyses reveal that our framework adapts its tool usage to specific task demands, and that verifier-guided replanning measurably improves both objective evidence support and verifier confidence. These findings advocate for a paradigm shift: mastering ancient documents is not solely an LVLM scaling problem, but fundamentally an evidence-grounding and reliability challenge.

Our contributions are summarized as follows:
\begin{itemize}
    \item We reformulate Chinese ancient document understanding as an evidence-grounded inference problem beyond direct LVLM answering, where systems expose supporting evidence, verification reports, abstention decisions, and execution traces in addition to final answers.

    \item We introduce \textsc{SAGE}, an evidence-grounded multi-agent framework that coordinates task-aware planning, tool-mediated evidence acquisition, claim-level verification, bounded replanning, and traceable diagnosis through role-specialized agents and a constrained shared-state runtime.

    \item We comprehensively evaluate \textsc{SAGE} on the AncientDoc benchmark across three backbone models. The results demonstrate consistent improvements over matched direct-answering baselines, while ablation and trace analyses validate the roles of planning, tool-mediated execution, and scholarly verification.
\end{itemize}
\section{Related Work}

\subsection{Document and Ancient-Document Understanding Benchmarks}

Document understanding has been extensively studied through document visual question answering, OCR-centric evaluation, and long-context multimodal benchmarks. Representative benchmarks such as DocVQA, TextVQA, OCRBench, OCRBench v2, and MMLongBench-Doc evaluate models' abilities to read, localize, and reason over textual or visual elements in document images and long documents~\citep{mathew2020docvqa,singh2019vqamodelsread,liu2024ocrbench,fu2024ocrbench,NEURIPS2024_ae0e4328}. These benchmarks have substantially advanced modern document understanding, but they mainly focus on contemporary documents, scene text, or general multimodal inputs.

Chinese ancient documents introduce a more specialized setting. They involve vertical layouts, traditional and variant characters, dense annotations, missing punctuation, classical Chinese expressions, implicit semantics, and historical-cultural references. Recent benchmarks have begun to evaluate models on Classical Chinese and ancient-document tasks. Text-based benchmarks such as AC-EVAL, C$^3$Bench, Fùxì, and WenyanBENCH focus on ancient Chinese language understanding, translation, knowledge, and generation~\citep{wei-etal-2024-ac,cao2024c3benchcomprehensiveclassicalchinese,zhao2025fuxibenchmarkevaluatinglanguage,yao2025wenyangptlargelanguagemodel}. Historical document resources and multimodal benchmarks such as HisDoc1B, BABMLLM, and AncientDoc further extend evaluation to ancient document images and ancient books~\citep{shi2025large,yu2025benchmarkingvisionlanguagemodelschinese}. Most directly related to our work, AncientDoc evaluates LVLMs on Chinese ancient document understanding across page-level OCR, vernacular translation, reasoning-based QA, knowledge-based QA, and linguistic variant QA.

These benchmarks demonstrate that ancient Chinese and ancient-document understanding require specialized evaluation beyond modern document tasks. However, they primarily evaluate final model outputs under a direct-answering paradigm. In contrast, \textsc{SAGE} focuses on the inference process itself: how an ancient-document system can acquire evidence, generate tentative answers, verify claims, abstain under insufficient evidence, and provide trace-based diagnostics during inference.

\subsection{Ancient Chinese Models and Historical Document Processing}

Another line of work develops domain-specific models and methods for ancient Chinese processing. Early pretrained models such as AnchiBERT adapt representation learning to ancient Chinese corpora, while recent LLM-based systems such as WenyanGPT and TongGu further explore continued pretraining, instruction tuning, retrieval augmentation, and knowledge-grounded Classical Chinese understanding~\citep{tian2021anchibertpretrainedmodelancient,yao2025wenyangptlargelanguagemodel,cao-etal-2024-tonggu}.

\textsc{SAGE} is complementary to these model-centric and translation-centric approaches. Rather than training a new ancient Chinese model, \textsc{SAGE} is designed as a model-agnostic inference-time framework for system-level evidence grounding. Rather than targeting text-only Classical Chinese understanding or translation alone, it addresses multi-task Chinese ancient document understanding from page images, where visual reading, text normalization, tool-mediated evidence acquisition, answer generation, and verification must be coordinated. Its core contribution is to operationalize ancient-document understanding as an evidence-grounded inference process, with explicit support for claim verification, bounded replanning, abstention, and trace-based component analysis.
\section{SAGE Framework}

\begin{figure*}[t]
    \centering
    \includegraphics[width=0.95\textwidth]{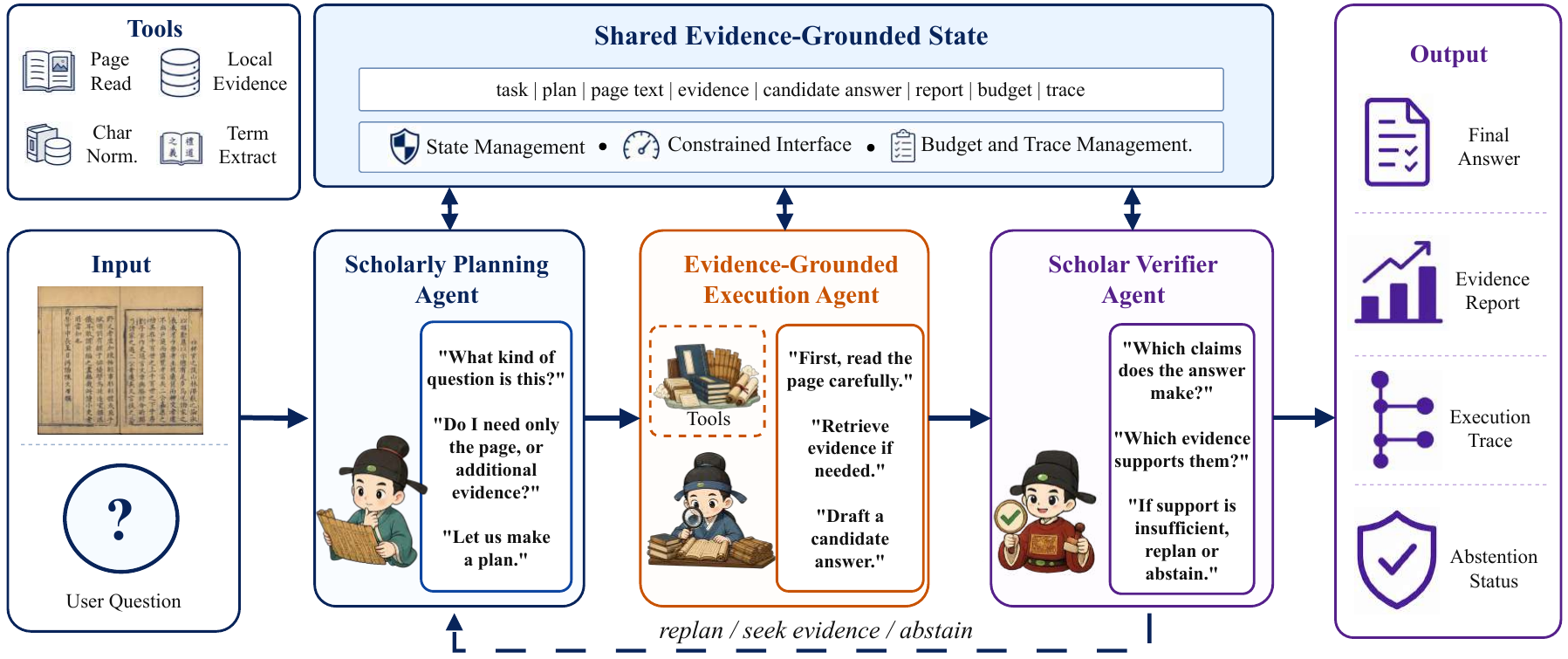}
\caption{
Overview of \textbf{SAGE}. SAGE reformulates Chinese ancient document understanding from direct answering into evidence-grounded inference by coordinating planning, tool-augmented execution, and verification under a constrained shared-state runtime. The runtime maintains an evidence-grounded state, exposes a constrained action interface, tracks budget, and records execution traces.
}
    \label{fig:sage-framework}
\end{figure*}

\subsection{Overview}

As shown in Figure~\ref{fig:sage-framework}, \textsc{SAGE} is an evidence-grounded multi-agent framework that organizes Chinese ancient document understanding as a stateful evidence-seeking process. Given an ancient document image $I$ and a natural-language question $Q$, the system maintains a shared evidence-grounded state and coordinates three role-specialized agent modules: a \emph{Scholarly Planning Agent}, a \emph{Tool-Augmented Execution Agent}, and a \emph{Scholar Verifier Agent}. These modules are not executed as an unconstrained conversation or a fixed cascade. Instead, they operate under a constrained shared-state runtime, which regulates the available actions, tracks evidence and budget, records execution traces, and applies predefined control rules over the shared state to stop, replan, or abstain when necessary.

For each input, \textbf{SAGE} outputs
\begin{equation}
O = (A, E, R, z, \tau),
\end{equation}
where $A$ is the final answer, $E$ is the collected evidence, $R$ is an evidence report, $z$ is a binary abstention indicator, and $\tau$ is the execution trace. These auxiliary outputs are not used to replace final-answer evaluation, but to make the answer-generation process more grounded, controllable, and diagnosable.

\subsection{Scholarly Planning Agent}

The Scholarly Planning Agent maps the input question to an executable evidence-seeking plan. Given the question $Q$ and an initial budget $b_0$, the planner predicts the task type, identifies the required grounding sources, and selects a sequence of actions from the runtime-defined action space. Formally, the initial plan is:
\begin{equation}
P_0 = \mathrm{Plan}(Q, b_0) = (y, \mathbf{a}_0, \rho_0, b_0),
\end{equation}
where $y$ is the predicted task type, $\mathbf{a}_0=(a_1,\ldots,a_m)$ is the planned action sequence, $\rho_0$ specifies the evidence requirement and verification criterion, and $b_0$ denotes the initial budget, such as the maximum number of tool calls or replanning rounds.

The plan determines what evidence should be acquired before answer synthesis. For example, translation-oriented questions mainly require page reading and text normalization, knowledge-intensive questions require concept extraction and source-constrained retrieval. The planning agent therefore produces a strategy rather than a final answer, separating task decomposition from evidence acquisition and answer generation. Typical task-specific planning strategies are summarized in Table~\ref{tab:task-strategy}.

When the runtime later triggers a bounded replanning round, the same planner updates the action plan based on the current evidence state and remaining budget, allowing the system to adapt its evidence-seeking behavior rather than following a fixed pipeline.

\begin{table}[t]
\centering
\scriptsize
\setlength{\tabcolsep}{2.2pt}
\renewcommand{\arraystretch}{1.15}
\begin{tabular}{p{0.18\linewidth}p{0.27\linewidth}p{0.43\linewidth}}
\toprule
\textbf{Task} & \textbf{Evidence Need} & \textbf{Planned Strategy} \\
\midrule
OCR 
& Visual page evidence 
& Read page text and check recognition validity. \\
\midrule
Translation 
& Page / normalized text 
& Read and normalize text, then generate a page-grounded vernacular translation. \\
\midrule
Knowledge QA 
& Concept / historical evidence 
& Extract key concepts, retrieve source-constrained evidence, and verify factual claims. \\
\midrule
Reasoning QA 
& Page-grounded evidence 
& Use recognized page text and question constraints to perform document-grounded reasoning. \\
\midrule
Linguistic QA 
& Text + style evidence 
& Use page text with variant, rhetoric, style, or genre evidence for linguistic analysis. \\
\bottomrule
\end{tabular}
\caption{Task-aware planning strategies in \textbf{SAGE}. The planner selects evidence requirements and action strategies before answer synthesis.}
\label{tab:task-strategy}
\end{table}

\subsection{Evidence-Grounded Execution Agent}

The Evidence-Grounded Execution Agent instantiates the current plan into concrete tool-mediated actions. Given the document image $I$, question $Q$, current plan $P_k$, and shared state $s_k$, the agent executes only actions allowed by the constrained runtime, updates the evidence state, and produces a candidate answer for later verification.

The execution agent operates over a constrained action space: \textit{read}, \textit{normalize}, \textit{extract}, \textit{retrieve}, and \textit{synthesize}. The \textit{read} action applies OCR or page-reading tools to obtain textual evidence from the document image. The \textit{normalize} action produces analyzable normalized forms while preserving the original recognized text, including traditional-to-simplified conversion and variant-character normalization when applicable. The \textit{extract} action identifies candidate concepts, entities, book titles, names, places, or classical expressions from the question and page text. The \textit{retrieve} action is invoked primarily for knowledge-intensive questions and uses the page context to search a frozen local knowledge evidence pool for relevant concept-level or historical-cultural evidence. The \textit{synthesize} action generates a candidate answer conditioned on the current page text and evidence state. More details about the tool implementations and invocation protocols are provided in Appendix~\ref{app:tools_retrieval}.

Formally, the execution agent updates the page text and evidence state through:
\begin{equation}
(X_k, E_k) = \mathrm{Exec}(I, Q, P_k, s_k),
\end{equation}
where $X_k$ denotes the recognized or normalized page text and $E_k$ denotes the accumulated evidence state. The evidence state includes page-derived evidence for all tasks and may additionally include retrieved evidence for knowledge-intensive questions. Retrieved evidence is not treated as an answer source by itself; instead, it provides contextual support for answer synthesis and later verification.

Given the page text and accumulated evidence, the execution agent generates a candidate answer:
\begin{equation}
\hat{A}_k = \mathrm{Solve}(Q, X_k, E_k, P_k),
\end{equation}
where $\hat{A}_k$ is treated as a hypothesis to be verified rather than the final output.

\subsection{Scholar Verifier Agent}

The Scholar Verifier Agent evaluates whether a candidate answer is sufficiently supported by the available page content and evidence state. Given the question $Q$, page text $X_k$, collected evidence $E_k$, and candidate answer $\hat{A}_k$, the verifier produces an evidence report $R_k$, which is stored in the shared state for runtime-level control.

For QA-oriented tasks, the verifier first decomposes the candidate answer into atomic claims:
\begin{equation}
C_k = \{c_1, c_2, \ldots, c_n\}.
\end{equation}
It then estimates the support relation between each claim and the available grounding context:
\begin{equation}
v_i = \mathrm{Verify}(c_i, Q, X_k, E_k), \quad v_i \in \mathcal{Y},
\end{equation}
where $\mathcal{Y}=\{\textsc{Sup}, \textsc{Con}, \textsc{Ins}, \textsc{N/A}\}$ denotes supported, contradicted, insufficient, and not applicable, respectively. The claim-level verification results are denoted as:
\begin{equation}
V_k = \{v_1, v_2, \ldots, v_n\}.
\end{equation}

The verifier aggregates these judgments into an evidence report:
\begin{equation}
R_k = \mathrm{Aggregate}(C_k, V_k, X_k, E_k),
\end{equation}
which summarizes evidence coverage, unsupported and contradicted claim rates, support confidence, and an abstention recommendation. If the main claims are sufficiently supported, the candidate answer becomes eligible for finalization by the runtime. Otherwise, the verifier records whether the current answer lacks support, misses necessary evidence, or should be rejected. The constrained shared-state runtime stores this feedback and may trigger replanning or abstention under the remaining budget.

The verifier applies task-sensitive criteria. Knowledge-intensive questions require stricter claim-evidence grounding and may trigger additional source-constrained retrieval or abstention. For translation, reasoning, and linguistic analysis, where the page text is often the primary grounding source, verification focuses on output validity, consistency with the document content, adequacy, and unsupported claims rather than exhaustive external support.

\begin{table*}[t]
\centering
\small
\setlength{\tabcolsep}{5pt}
\renewcommand{\arraystretch}{1.15}
\resizebox{\textwidth}{!}{
\begin{tabular}{llcccccccc}
\toprule
\multirow{2}{*}{\textbf{Model / Backbone}}
& \multirow{2}{*}{\textbf{Paradigm}}
& \multicolumn{2}{c}{\textbf{Translation}}
& \multicolumn{2}{c}{\textbf{Reasoning QA}}
& \multicolumn{2}{c}{\textbf{Knowledge QA}}
& \multicolumn{2}{c}{\textbf{Linguistic QA}} \\
\cmidrule(lr){3-4}
\cmidrule(lr){5-6}
\cmidrule(lr){7-8}
\cmidrule(lr){9-10}
&
& \textbf{CHRF++}
& \textbf{BS-F1}
& \textbf{CHRF++}
& \textbf{BS-F1}
& \textbf{CHRF++}
& \textbf{BS-F1}
& \textbf{CHRF++}
& \textbf{BS-F1} \\
\midrule
DeepSeek-VL2~\cite{wu2024deepseek}
& Direct
& 0.49 & 50.27 & 3.42 & 59.09 & 3.77 & 59.83 & 3.70 & 60.05 \\
LLaVA-OneVision-72B~\cite{li2024llava}
& Direct
& 0.44 & 48.83 & 5.96 & 67.20 & 5.56 & 66.31 & 2.51 & 57.57 \\
InternVL3-78B~\cite{zhu2025internvl3}
& Direct
& 4.45 & 62.40 & 4.95 & 65.99 & 5.25 & 65.79 & 2.38 & 57.78 \\
Qwen2.5-VL-72B~\cite{qwen2.5-VL}
& Direct
& 9.77 & 69.87 & \underline{9.04} & \textbf{71.40} & 7.82 & 69.15 & 3.65 & 59.34 \\
\midrule
GPT-4o~\cite{hurst2024gpt}
& Direct
& 3.02 & 58.86 & 7.99 & 70.52 & 8.02 & 70.01 & 4.16 & \underline{64.58} \\
Gemini2.5-Pro~\cite{comanici2025gemini}
& Direct
& 11.41 & \underline{72.50} & 8.68 & 69.33 & 8.88 & 68.94 & 5.22 & 62.06 \\
Doubao-V2~\cite{seed2025seed1_5vl}
& Direct
& 0.56 & 52.39 & 7.15 & 68.78 & 8.75 & 69.15 & 3.32 & 57.70 \\
Qwen-VL-Max~\cite{Qwen-VL}
& Direct
& \underline{12.30} & 71.03 & 8.60 & \underline{71.30} & 7.58 & 68.67 & 3.31 & 58.77 \\
\midrule
\multirow{2}{*}{InternVL3-8B~\cite{zhu2025internvl3}}
& Direct
& 0.85 & 53.16 & 5.21 & 65.62 & 4.75 & 63.60 & 1.93 & 55.96 \\
& \cellcolor{gray!12}SAGE
& \cellcolor{gray!12}4.91 
& \cellcolor{gray!12}61.89 
& \cellcolor{gray!12}8.98 
& \cellcolor{gray!12}69.00 
& \cellcolor{gray!12}\underline{10.84} 
& \cellcolor{gray!12}\underline{70.40} 
& \cellcolor{gray!12}\underline{6.02} 
& \cellcolor{gray!12}64.51 \\
\cmidrule(lr){1-10}
\multirow{2}{*}{Qwen2.5-VL-7B~\cite{qwen2.5-VL}}
& Direct
& 7.04 & 65.59 & 7.34 & 69.96 & 5.87 & 66.75 & 2.65 & 57.48 \\
& \cellcolor{gray!12}SAGE
& \cellcolor{gray!12}10.16 
& \cellcolor{gray!12}67.70 
& \cellcolor{gray!12}8.04 
& \cellcolor{gray!12}70.99 
& \cellcolor{gray!12}8.45 
& \cellcolor{gray!12}69.92 
& \cellcolor{gray!12}4.05 
& \cellcolor{gray!12}62.18 \\
\cmidrule(lr){1-10}
\multirow{2}{*}{Qwen3.5-9B~\cite{qwen3.5}}
& Direct
& 7.71 & 68.03 & 6.88 & 68.56 & 6.80 & 66.83 & 1.83 & 55.63 \\
& \cellcolor{gray!12}SAGE
& \cellcolor{gray!12}\textbf{16.01} 
& \cellcolor{gray!12}\textbf{75.29} 
& \cellcolor{gray!12}\textbf{9.35} 
& \cellcolor{gray!12}69.10 
& \cellcolor{gray!12}\textbf{11.81} 
& \cellcolor{gray!12}\textbf{71.12} 
& \cellcolor{gray!12}\textbf{7.72} 
& \cellcolor{gray!12}\textbf{65.35} \\
\bottomrule
\end{tabular}
}
\caption{
Main results on AncientDoc. OCR is excluded because it primarily depends on external page-reading tools. We report CHRF++ and BERTScore F1 for four understanding tasks. Direct baselines use task-specific prompts with known task types, whereas gray rows denote SAGE runs without task-type labels. Best and second-best results are in bold and underline.
}
\label{tab:main-results}
\end{table*}

\subsection{Constrained Shared-State Runtime}

The constrained shared-state runtime provides the system-level infrastructure that organizes the three agent modules into a coherent evidence-grounded inference process. It is not an additional autonomous agent; rather, it defines the shared evidence state, constrained tool interface, budget limits, and trace schema used throughout inference. At iteration $k$, the runtime maintains:
\begin{equation}
s_k = (y, P_k, X_k, E_k, \hat{A}_k, R_k, b_k, \tau_k),
\end{equation}
where $y$ is the predicted task type, $P_k$ is the current plan, $X_k$ is the recognized or normalized page text, $E_k$ is the accumulated evidence state, $\hat{A}_k$ is the current candidate answer, $R_k$ is the evidence report, $b_k$ is the remaining budget, and $\tau_k$ is the execution trace.

Unlike a fixed pipeline, the runtime does not require every input to follow the same action sequence. It executes the current plan under the constrained tool interface, updates the shared state after each tool call and verification step, and applies predefined control rules over the updated state:
\begin{equation}
d_k = \mathrm{Rule}(s_k), \quad d_k \in \mathcal{D},
\end{equation}
where
\begin{equation}
\mathcal{D}=\{\textsc{Stop}, \textsc{Replan}, \textsc{Abstain}\}.
\end{equation}
In our implementation, \textsc{Stop} is selected when the evidence report indicates sufficient support or the task-specific adequacy check passes. \textsc{Replan} is selected when the current state suggests missing evidence and the remaining budget allows another evidence-seeking round. \textsc{Abstain} is selected when evidence remains insufficient, the output fails verification, or the budget is exhausted.

The runtime follows three design principles.

\paragraph{Shared state management.}
The runtime maintains a unified state across planning, execution, and verification, including the predicted task type, current plan, page text, evidence, candidate answer, evidence report, remaining budget, and execution trace. This allows modules to reuse accumulated evidence and intermediate results instead of restarting from scratch.

\paragraph{Constrained action interface.}
The runtime exposes a fixed set of actions, including page reading, text normalization, concept extraction, local evidence retrieval, answer synthesis, and verification. Agent modules can only interact with the system through this interface, which prevents hallucinated tool use and keeps inference reproducible.

\paragraph{Budget and trace management.}
The runtime enforces limits on tool calls and replanning rounds, and records the execution trajectory, including planned actions, tool calls, retrieved evidence, candidate answers, evidence reports, replanning decisions, abstention decisions, and cost information when available. These traces support component-level diagnosis beyond final-answer accuracy.

In summary, the runtime turns planning, execution, and verification into a stateful and traceable evidence-grounded process. It provides shared memory, constrained tool access, budget control, and diagnostic logging, while keeping inference more adaptive than a fixed one-pass pipeline.
\section{Experiments}

We evaluate whether \textsc{SAGE} improves Chinese ancient document understanding beyond direct LVLM answering, and whether its evidence-grounded runtime produces more grounded and diagnosable outputs. Specifically, we study three questions:
\textbf{(1)} Does \textsc{SAGE} improve final-answer quality over matched direct-answering baselines?
\textbf{(2)} Are the gains consistent across different backbone models?
\textbf{(3)} How do planning, tool-mediated evidence acquisition, scholarly verification, and bounded evidence seeking contribute to performance and grounding?

\subsection{Experimental Setup}

\paragraph{Benchmark and Metrics.}
We evaluate \textsc{SAGE} on AncientDoc~\cite{yu2025benchmarkingvisionlanguagemodelschinese}, which contains five task types: OCR, translation, reasoning-based QA, knowledge-based QA, and linguistic QA. Our main evaluation focuses on the four understanding-oriented tasks: translation, reasoning-based QA, knowledge-based QA, and linguistic QA. We exclude OCR from the main comparison because OCR primarily measures page-reading or OCR tool quality, whereas our focus is on evidence acquisition, verification, and runtime control beyond raw text recognition. Following the AncientDoc evaluation protocol, we report CHRF++ and BERTScore F1 (BS-F1).

\paragraph{Models and Settings.}
We compare \textsc{SAGE} with direct-answering LVLM baselines. For matched comparisons, we instantiate \textsc{SAGE} with InternVL3-8B~\cite{zhu2025internvl3}, Qwen2.5-VL-7B~\cite{qwen2.5-VL}, and Qwen3.5-9B~\cite{qwen3.5}, and compare each against its direct-answering counterpart. Direct baselines generate answers from the image-question input in a single pass using task-specific prompts, while \textsc{SAGE} receives the image-question input and performs task prediction, evidence acquisition, verification, and bounded replanning through the constrained shared-state runtime. Unless otherwise specified, all \textsc{SAGE} variants use the same constrained tool interface, frozen local evidence pool, verification protocol, and budget constraints. More details about baseline prompts, tool implementations, retrieval settings, and budgets are provided in Appendix~\ref{app:implementation}.

\subsection{Main Results}

Table~\ref{tab:main-results} reports the main results on the four understanding-oriented AncientDoc tasks. The upper blocks list representative open-source and closed-source LVLMs under the direct-answering paradigm, while the lower block compares direct answering and \textsc{SAGE} under matched backbones: InternVL3-8B, Qwen2.5-VL-7B, and Qwen3.5-9B. For all tasks, we report CHRF++ and BERTScore F1 (BS-F1).

\textsc{SAGE} consistently improves over matched direct-answering baselines. Across three backbones and eight task metrics, \textsc{SAGE} improves every corresponding direct baseline, showing that the gains are not tied to a single model family. The improvements are especially large on Knowledge QA and Linguistic QA, where explicit evidence organization and verification are most beneficial. For example, with InternVL3-8B, \textsc{SAGE} improves Knowledge QA by +6.09 CHRF++ and +6.80 BS-F1, and Linguistic QA by +4.09 CHRF++ and +8.55 BS-F1.

Compared with much larger direct-answering LVLMs, \textsc{SAGE} also remains competitive. Qwen3.5-9B with \textsc{SAGE} achieves the best score on seven out of eight metrics and surpasses all direct LVLM baselines on Translation, Knowledge QA, and Linguistic QA. The only exception is Reasoning QA BS-F1, where Qwen2.5-VL-72B Direct is slightly higher. These results suggest that evidence-grounded inference and verification can improve ancient-document understanding beyond relying only on larger base LVLMs.

\begin{figure*}[t]
    \centering
    \includegraphics[width=\textwidth]{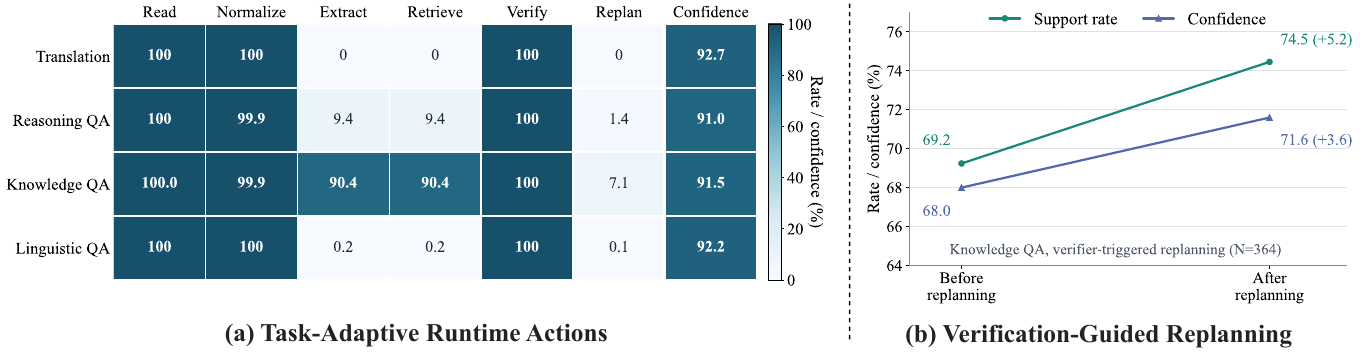}
    \caption{
    Trace-based behavior analysis of \textsc{SAGE}. 
    (a) \textit{Task-Adaptive Runtime Actions}: action patterns across task types. Each cell reports the percentage of examples invoking the corresponding action, while the last column reports average final verifier confidence.
    (b) \textit{Verification-Guided Replanning}: analysis on Knowledge QA examples where replanning is triggered. After replanning, both trace-derived support rate and verifier confidence increase.
    }
    \label{fig:trace-behavior}
\end{figure*}

\subsection{Ablation Study}

We conduct ablation studies on reasoning-based QA and knowledge-based QA, which test page-grounded reasoning and knowledge-intensive evidence use. Since the constrained shared-state runtime and the Evidence-Grounded Execution Agent define the shared state and tool-mediated execution process, we keep them fixed and ablate replaceable components: planning, page reading, local retrieval, and verification. All ablations use Qwen3.5-9B as the backbone unless otherwise specified.

For \textit{w/o Planner}, we replace the Scholarly Planning Agent with a fixed task-agnostic action sequence. For \textit{w/o Page Reader}, we disable the explicit page-reading tool and rely on the backbone LVLM to read the document image directly. For \textit{w/o Retrieval}, we disable retrieval from the frozen local evidence pool, which is only reported for Knowledge QA. For \textit{w/o Verifier}, the candidate answer is returned without an evidence report or claim-level support judgments.

\begin{table}[t]
\centering
\small
\setlength{\tabcolsep}{3.5pt}
\renewcommand{\arraystretch}{1.12}
\begin{tabular}{lcc|cc}
\toprule
\multirow{2}{*}{\textbf{Variant}}
& \multicolumn{2}{c|}{\textbf{Reasoning QA}}
& \multicolumn{2}{c}{\textbf{Knowledge QA}} \\
\cmidrule(lr){2-3}
\cmidrule(lr){4-5}
& \textbf{CHRF++}
& \textbf{BS-F1}
& \textbf{CHRF++}
& \textbf{BS-F1} \\
\midrule
Direct LVLM
& 6.88 & 68.56 & 6.8 & 66.83 \\
w/o Planner
& 9.27 & 67.06 & 9.34 & 67.8 \\
w/o Page Reader
& 9.15 & 68.65 & 11.58 & 70.79 \\
w/o Retrieval
& - & - & 11.78 & 70.89 \\
w/o Verifier
& 9.13 & 68.98 & 11.23 & 70.08 \\
\textbf{SAGE}
& 9.35 & 69.10 & 11.81 & 71.12 \\
\bottomrule
\end{tabular}
\caption{
Ablation study on reasoning-based QA and knowledge-based QA. We report CHRF++ and BERTScore F1 (BS-F1). Direct LVLM denotes single-pass answering with the same backbone. Other variants remove one replaceable component from \textsc{SAGE} while keeping the backbone, constrained runtime, execution agent, and budget settings fixed. Retrieval is ablated only on Knowledge QA, where the local evidence pool is used.
}
\label{tab:ablation}
\end{table}

As shown in Table~\ref{tab:ablation}, the full \textsc{SAGE} achieves the best results on both tasks. All \textsc{SAGE} variants outperform the Direct LVLM baseline, indicating that the shared evidence-grounded runtime already provides substantial gains over single-pass answering. Removing the planner or verifier leads to consistent drops, especially in BS-F1, showing the importance of task-aware planning and scholarly verification. The effect of local retrieval is positive but modest on Knowledge QA, suggesting that \textsc{SAGE}'s main gains come from page-grounded evidence organization, tool-mediated execution, and verification rather than from external knowledge alone.

\subsection{Trace-based Behavior Analysis}

Beyond final answer scores, \textsc{SAGE} records execution traces for each example, including planned actions, invoked tools, retrieved evidence, verification reports, and replanning decisions. We use these traces to examine whether \textsc{SAGE} produces task-adaptive execution patterns and whether verifier feedback leads to measurable changes in the inference process.

Figure~\ref{fig:trace-behavior}(a) shows the action patterns across the four understanding-oriented tasks. Page reading, normalization, and verification are invoked for almost all examples, forming the common evidence-grounded runtime. In contrast, extraction and retrieval exhibit clear task specificity. Knowledge QA invokes extraction and local evidence retrieval in $90.39\%$ of the examples, while reasoning QA invokes them in only $9.42\%$ of the examples. Translation and linguistic QA rarely trigger retrieval, indicating that they mainly rely on page-level evidence rather than external knowledge. This supports that \textsc{SAGE} does not simply execute a fixed pipeline, but selects different evidence paths according to task semantics. The final confidence column further shows that the verifier maintains comparable confidence across tasks, suggesting that the adaptive use of evidence does not come at the cost of unstable verification behavior.

Figure~\ref{fig:trace-behavior}(b) further analyzes the Knowledge QA examples where the verifier triggers replanning. After replanning, the trace-derived support rate increases from $69.23\%$ to $74.45\%$, and verifier confidence increases from $67.99\%$ to $71.59\%$. These results suggest that verifier feedback helps \textsc{SAGE} produce more strongly supported answers in cases initially judged as insufficiently grounded.

It should be noted that these statistics are derived from the system's own verification reports rather than human factuality annotations. Therefore, they are used as behavioral diagnostics of the runtime: the traces show when different tools are invoked, when replanning is triggered, and how the verifier's evidence-support assessment changes after an additional inference round.
\section{Conclusion}

We presented \textbf{SAGE}, an evidence-grounded multi-agent framework that moves Chinese ancient document understanding from direct LVLM answering toward evidence-grounded inference. SAGE coordinates role-specialized scholarly agents under a constrained shared-state runtime, supporting task-aware planning, tool-mediated evidence acquisition, scholarly verification, and bounded evidence seeking. By making evidence acquisition and verification explicit, SAGE provides a traceable basis for diagnosing unsupported or insufficiently grounded answers. Experiments show that SAGE improves over matched direct-answering baselines across multiple backbones, especially on knowledge-intensive and linguistic tasks. Ablation and trace-based analyses indicate that these gains arise from structured page-grounded evidence use, tool-mediated execution, and verification rather than larger backbones or external knowledge alone. Overall, ancient document understanding should be treated not only as a model-scaling problem, but also as an evidence-grounding and reliability challenge.

\section*{Limitations}

SAGE is an initial exploration of evidence-grounded Chinese ancient document
understanding. Although it improves over matched direct-answering baselines on
AncientDoc, the current evaluation is limited to the tasks and data distribution
covered by this benchmark, and its applicability to broader historical
collections, different document layouts, and other ancient languages remains to
be validated. In addition, the verifier and trace-based support statistics are
system-generated diagnostics rather than human factuality annotations, so they
should be interpreted as indicators of runtime behavior rather than definitive
proof of answer correctness. Finally, SAGE introduces additional inference
cost through planning, tool use, verification, and replanning. Future work may
explore stronger evidence sources, human-grounded verification, and more
efficient execution.

\bibliography{main}

\appendix

\section{Implementation Details}
\label{app:implementation}

This section provides implementation details for the experimental setup in the
main paper, including model backbones, direct-answering baselines, SAGE
runtime configuration, decoding settings, and execution budgets.

\subsection{Backbone Models}
\label{app:backbones}

We instantiate SAGE with three LVLM backbones: InternVL3-8B,
Qwen2.5-VL-7B, and Qwen3.5-9B. For each backbone, we compare SAGE with a
matched direct-answering baseline using the same underlying model. This matched
comparison isolates the effect of the evidence-grounded runtime from the effect
of backbone choice. Unless otherwise specified, all backbone parameters are
kept fixed during evaluation; SAGE changes only the inference procedure by
adding planning, tool-mediated evidence acquisition, verification, and bounded
replanning.

\subsection{Direct-Answering Baselines}
\label{app:direct_baselines}

The direct-answering baseline receives the document image and the user question
as input and generates the final answer in a single pass. We use task-specific
prompts for the four understanding-oriented tasks evaluated in the main paper:
translation, reasoning-based QA, knowledge-based QA, and linguistic QA. The
baseline does not receive retrieved evidence, intermediate verification
reports, or replanning feedback. This setting represents the standard LVLM
usage pattern where the model is expected to read, reason, and answer directly
from the image-question input.

\subsection{SAGE Runtime Configuration}
\label{app:runtime_config}

SAGE uses the same input as the direct-answering baseline but executes an
evidence-grounded runtime before producing the final answer. The runtime consists
of four main steps:
(i) task prediction, which identifies the type of ancient-document understanding
task;
(ii) evidence acquisition, which invokes tools such as page reading,
normalization, extraction, and local evidence retrieval when applicable;
(iii) answer generation, which produces a candidate answer grounded in the
collected evidence; and
(iv) scholarly verification, which checks whether the answer is supported by
the available evidence and may trigger bounded replanning.

All SAGE variants use the same constrained tool interface, frozen local
evidence pool, verification protocol, and budget constraints. Ablation variants
remove or replace a specific component while keeping the remaining runtime
configuration fixed.

\subsection{Decoding Settings}
\label{app:decoding}

For fair comparison, we use the same decoding configuration for direct answering
and SAGE answer generation under the same backbone. We use deterministic or
low-temperature decoding to reduce variance in final answers. The maximum
generation length is set to cover the longest expected outputs in translation
and knowledge-intensive QA tasks. The same decoding settings are also used for
planning and verification unless otherwise specified.

\subsection{Execution Budgets}
\label{app:budgets}

SAGE operates under bounded inference budgets to prevent unbounded tool use
or indefinite replanning. We limit the maximum number of tool calls, retrieved
evidence snippets, verification passes, and replanning rounds for each example.
The same budgets are used across all backbone models and tasks. Replanning is
triggered only when the verifier judges that the current answer has insufficient
evidence support or low confidence. If the budget is exhausted, the system
returns the best supported answer produced so far.

\subsection{Reproducibility Notes}
\label{app:reproducibility}

All experiments are conducted under the official AncientDoc evaluation protocol.
We report CHRF++ and BERTScore F1 for the four understanding-oriented tasks:
translation, reasoning-based QA, knowledge-based QA, and linguistic QA. OCR is
excluded from the main comparison because the focus of SAGE is evidence
acquisition, verification, and runtime control beyond raw page reading. We keep
the evidence pool frozen during evaluation and apply the same retrieval and
verification settings to all matched SAGE variants.

\section{Prompts}
\label{sec:appendix-prompts}

This appendix reports the core prompt templates used by SAGE.
The implementation prompts are written in Chinese because the target
documents and tasks are Chinese ancient-document understanding tasks.
For readability, we present lightly translated templates while keeping
the task labels, tool names, and runtime placeholders unchanged.

\subsection{Prompt Organization}

SAGE uses a small family of prompts rather than a single monolithic
instruction. The task classifier first maps the user request to one of
five task types. The scholarly planning prompt then produces a structured
tool plan under the current budget and evidence state. Task-specific
answer prompts synthesize candidate outputs from page text and, when
needed, local evidence. Finally, the verifier prompt checks whether the
candidate answer is supported by the available evidence and returns a
structured report used by the runtime to stop, replan, or abstain.

\subsection{Task Classification Prompt}

\noindent\textbf{Purpose.}
The classifier normalizes heterogeneous user questions into the fixed
task ontology used by the runtime.

\begin{quote}\small
You are the task classifier for SAGE. Given the user question,
decide which of the following five task types it belongs to:
\texttt{ocr}, \texttt{translation}, \texttt{reasoning\_qa},
\texttt{knowledge\_qa}, or \texttt{linguistic\_qa}.

\texttt{ocr}: the user asks to extract, recognize, transcribe, or copy
text from an image.

\texttt{translation}: the user asks to translate ancient text into modern
Chinese or vernacular Chinese.

\texttt{reasoning\_qa}: the user asks a question requiring causal,
semantic, ideological, or implicit inference based on page content.

\texttt{knowledge\_qa}: the user asks about people, places, allusions,
terms, historical facts, or cultural knowledge.

\texttt{linguistic\_qa}: the user asks about rhetoric, genre, language
style, period style, or literary features.

Output exactly one label and do not explain.

User question: \texttt{\{question\}}
\end{quote}

\subsection{Scholarly Planning Prompt}

\noindent\textbf{Purpose.}
The planner does not answer the question directly. It produces a JSON
strategy that can only call tools exposed by the constrained runtime.

\begin{quote}\small
You are the Scholarly Planning Agent of SAGE. You do not directly
answer the user question. Your only responsibility is to generate a
structured JSON strategy plan, which will be executed by the constrained
SAGE runtime.

Inputs: planning mode \texttt{\{planning\_mode\}}, round index
\texttt{\{round\_index\}}, task type \texttt{\{task\}}, user question
\texttt{\{question\}}, tool registry \texttt{\{tool\_registry\}}, task
budgets \texttt{\{budgets\}}, current state summary
\texttt{\{current\_state\_summary\}}, previous strategy
\texttt{\{previous\_strategy\}}, verification feedback
\texttt{\{verification\_feedback\}}, successful strategy memory
\texttt{\{strategy\_memory\}}, and failure memory
\texttt{\{failure\_memory\}}.

Planning principles:
Only use tools listed in the tool registry. Do not exceed the current
task budget. Do not generate or execute Python code. At the initial
planning stage.
\texttt{translation}, \texttt{reasoning\_qa}, \texttt{knowledge\_qa},
and \texttt{linguistic\_qa} usually require \texttt{normalize\_text}.
\texttt{knowledge\_qa} can use \texttt{retrieve\_evidence} over the
local knowledge pool. After \texttt{solve\_task}, always call
\texttt{verify\_answer}. During replanning, usually do not repeat OCR;
for \texttt{knowledge\_qa}, retrieve additional evidence if the verifier
requests it. If evidence remains insufficient, plan a conservative
verification-ending path that allows the verifier to trigger abstention.

Recommended initial workflows:
\texttt{ocr}: \texttt{ocr\_image} $\rightarrow$ \texttt{verify\_answer}.
\texttt{translation}: \texttt{ocr\_image} $\rightarrow$
\texttt{normalize\_text} $\rightarrow$ \texttt{solve\_task}
$\rightarrow$ \texttt{verify\_answer}.
\texttt{reasoning\_qa}: \texttt{ocr\_image} $\rightarrow$
\texttt{normalize\_text} $\rightarrow$ \texttt{solve\_task}
$\rightarrow$ \texttt{verify\_answer}.
\texttt{knowledge\_qa}: \texttt{ocr\_image} $\rightarrow$
\texttt{normalize\_text} $\rightarrow$ \texttt{extract\_terms}
$\rightarrow$ \texttt{retrieve\_evidence} $\rightarrow$
\texttt{solve\_task} $\rightarrow$ \texttt{verify\_answer}.
\texttt{linguistic\_qa}: \texttt{ocr\_image} $\rightarrow$
\texttt{normalize\_text} $\rightarrow$ \texttt{solve\_task}
$\rightarrow$ \texttt{verify\_answer}.

Return JSON only. Do not explain and do not use Markdown. The JSON must
contain the task label, maximum tool calls, an ordered list of steps, a
fallback list, and a task-specific output rule.
\end{quote}

\subsection{Page Reading Prompt}
\label{page_reading_prompt}

\noindent\textbf{Purpose.}
The page-reading prompt enforces document-layout assumptions for Chinese
ancient books and prevents explanatory text from entering the OCR state.

\begin{quote}\small
You are an OCR model specialized in Chinese ancient-book vertical pages.
Strictly recognize the text in the image. Read columns from right to left,
and read each column from top to bottom. Output only the OCR result; do
not output explanations, prefixes, suffixes, Markdown, numbering, or
prompting text. Preserve original characters and original wording. Do not
convert between simplified and traditional Chinese and do not modernize
the text. Do not guess missing characters from meaning. Main text,
interlinear notes, and small annotations should be output according to
their reading position if they belong to the textual content. Ignore
column lines, borders, stains, decorative patterns, seals, and non-textual
page numbers. The final output must contain only recognized text.
\end{quote}

\subsection{Task-Specific Answer Prompts}

\noindent\textbf{Purpose.}
The answer prompts share a common style constraint: answer directly,
avoid chain-of-thought or evidence narration, preserve key names and
terms, and abstain only when the available information is genuinely
insufficient.

\paragraph{Translation.}
\begin{quote}\small
Translate the ancient OCR text into modern vernacular Chinese. Inputs are
\texttt{\{ocr\_text\}} and normalized reference text
\texttt{\{normalized\_text\}}. Output only the translation. Do not explain
the translation process and do not start with phrases such as ``this
passage means''. Preserve the main information in the source. If the OCR
contains missing-character marks, handle them cautiously without excessive
inference. Perform semantic translation rather than character conversion;
rewrite classical syntax into natural modern Chinese, adding omitted
subjects, objects, or logical relations when necessary. Do not fabricate
information absent from the OCR text.
\end{quote}

\paragraph{Reasoning QA.}
\begin{quote}\small
Answer reasoning questions from the page text. Inputs are the question
\texttt{\{question\}}, ancient OCR or normalized text \texttt{\{text\}},
and available evidence \texttt{\{evidence\}}. Answer the question
directly without saying ``according to the page''. For causes, motives,
effects, or functions, use natural formulations such as ``because ...''
or ``the reason is ...''. For ideas, cultural meaning, or institutional
features, summarize with expressions such as ``this reflects ...''.
Preserve key terms, names, book titles, institutions, numbers, and proper
expressions from the question and page. Necessary semantic inference is
allowed, but the inference must be integrated into the answer rather than
shown as reasoning steps. Usually answer in one to three sentences.
\end{quote}

\paragraph{Knowledge QA.}
\begin{quote}\small
Answer knowledge questions about people, places, allusions, terms,
historical facts, or cultural knowledge using page text
\texttt{\{text\}} and local knowledge-base evidence \texttt{\{evidence\}}.
Give a concise answer similar to a dictionary-style explanation. Do not
separate page information from knowledge-base evidence and do not cite
the retrieval process. For terms, allusions, people, places, books, or
institutions, prefer forms such as ``X refers to ...'', ``X is ...'', or
``namely ...''. Include only the definition, source, function, influence,
or cultural meaning most relevant to the question. If evidence is limited
but the question concerns common knowledge, provide a cautious brief
answer; if it is impossible to judge, answer ``cannot be determined''.
\end{quote}

\paragraph{Linguistic QA.}
\begin{quote}\small
Answer questions about language style, rhetoric, genre, or literary
features using page text \texttt{\{text\}} and, when necessary, model
knowledge. Do not use web search. Provide a clear judgment. If asked
about rhetoric, first state the rhetorical category; if asked about
genre, first state the genre; if asked about language style, first
summarize the style. Point to concrete textual evidence such as words,
sentence patterns, parallel structures, or imagery. Avoid generic literary
comments detached from the page text. If the page text is insufficient,
say that the textual basis is insufficient.
\end{quote}

\subsection{Verifier Prompt}

\noindent\textbf{Purpose.}
The verifier checks candidate answers without rewriting the task from
scratch. It returns a structured report used by the runtime for stopping,
replanning, or abstention.

\begin{quote}\small
You are the Scholar Verifier Agent of SAGE. Your task is not to
answer the question again, but to perform a lightweight quality check on
the candidate answer. Adjust strictness by task type: knowledge QA and
reasoning QA require evidence consistency; translation and OCR only
require checks for format, completeness, and obvious errors.

Inputs: task type \texttt{\{task\}}, user question \texttt{\{question\}},
candidate answer \texttt{\{answer\}}, rule-split claims
\texttt{\{rule\_claims\}}, and available evidence \texttt{\{evidence\}}.

Use the labels \texttt{supported}, \texttt{contradicted},
\texttt{insufficient}, and \texttt{not\_applicable}. For
\texttt{knowledge\_qa}, prioritize retrieved evidence and do not support
claims without evidence. For \texttt{reasoning\_qa}, allow cautious
page-grounded inference. For \texttt{linguistic\_qa}, require concrete
textual or stylistic evidence. For \texttt{translation}, do not demand
external evidence; check whether the output is modern Chinese, preserves
the main information, and avoids irrelevant explanations or disclaimers.
For OCR, check whether usable OCR output exists.

Return JSON only. The JSON contains a cleaned final answer, claim list,
claim-level verifications, an evidence report, an abstention flag, a
confidence score, a \texttt{needs\_more\_evidence} flag, and an issue
list.
\end{quote}

\section{Tool Interface, Evidence Pool, and Retrieval}
\label{app:tools_retrieval}

This section describes the constrained evidence-acquisition interface used by
\textbf{SAGE}. We focus on four tools: page reading, character normalization,
term extraction, and local evidence retrieval. Answer synthesis is performed by
the Evidence-Grounded Execution Agent, and verification is handled by the
Scholar Verifier Agent; they are not counted as evidence-acquisition tools in
this section.

\subsection{Tool Registry}
\label{app:tool_registry}

SAGE exposes a small fixed registry of evidence-acquisition tools. Each tool
has a predefined input-output schema, and the runtime validates the planned
action sequence before execution. This prevents unconstrained tool use and makes
the evidence path traceable.

\begin{table}[t]
\centering
\small
\setlength{\tabcolsep}{3.5pt}
\renewcommand{\arraystretch}{1.12}
\begin{tabular}{p{2.0cm}p{2.1cm}p{2.4cm}}
\toprule
\textbf{Tool} & \textbf{Input} & \textbf{Output} \\
\midrule
Page Reader
& Document image
& Page-level OCR text and source metadata \\
Char Normalizer
& Raw OCR text
& Normalized text and character-level changes \\
Term Extractor
& Question and page text
& Candidate terms and retrieval queries \\
Local Evidence
& Question, terms, and page context
& Page evidence and local KB snippets \\
\bottomrule
\end{tabular}
\caption{
Evidence-acquisition tool interface used by SAGE. Answer synthesis and
verification are handled by the agent modules rather than listed as tools in
this table.
}
\label{tab:tool_registry}
\end{table}

\subsection{Page Reader}
\label{app:page_reader}

The Page Reader obtains page-grounded textual evidence from the document image.
In our experiments, it is implemented with Qwen-VL-OCR following the
page-reading prompt in Appendix~\ref{page_reading_prompt}. Given a document
image, the tool returns page-level OCR text together with basic source metadata.
The recognized text is used as the primary page-level evidence for all task
types and is passed to downstream normalization, extraction, answer synthesis,
and verification modules.

\begin{figure*}[t]
    \centering
    \includegraphics[width=\textwidth]{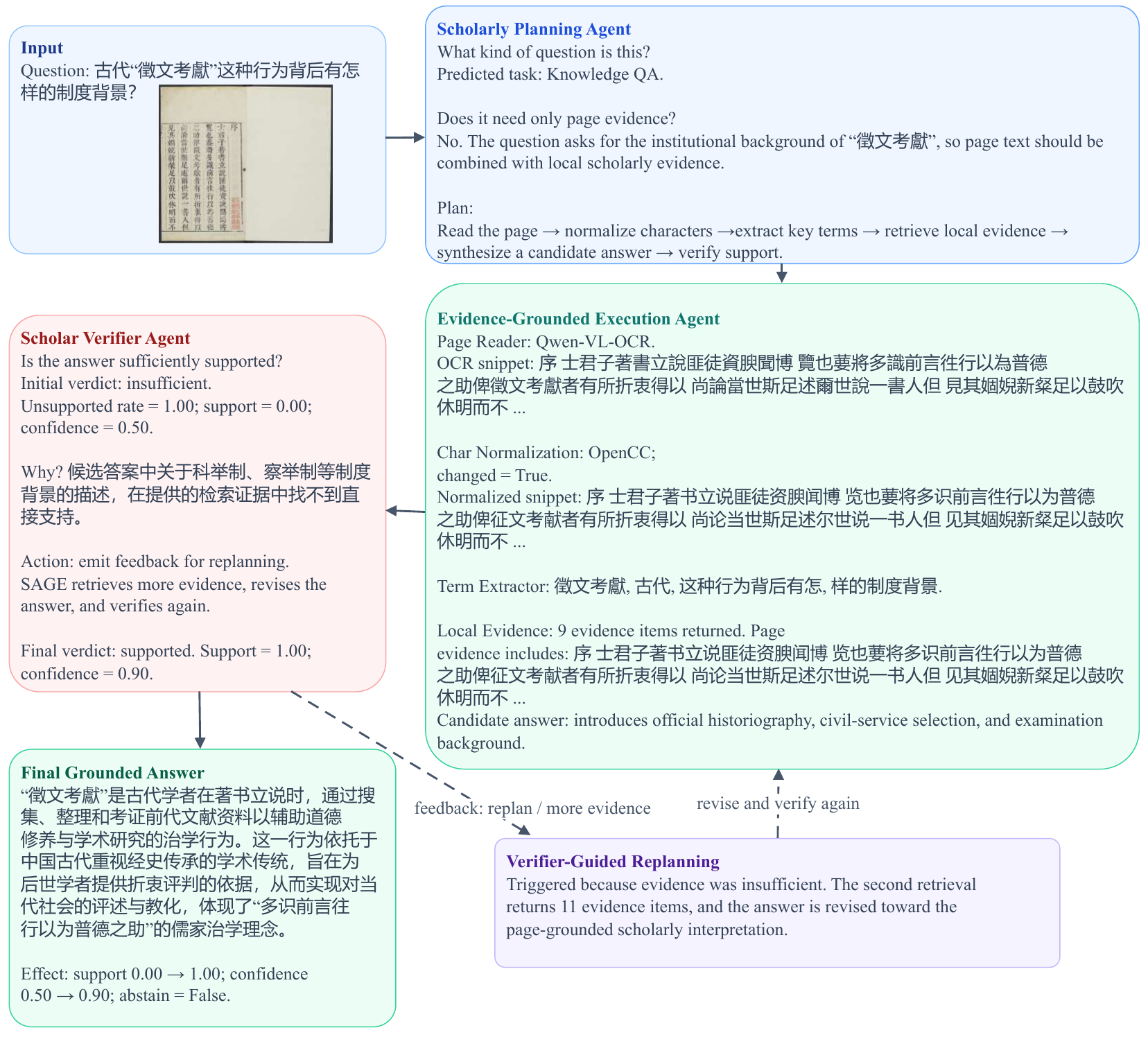}
    \caption{
    Qualitative trace of a Knowledge QA example. The verifier identifies
    unsupported institutional claims in the initial candidate answer, triggers
    replanning, and leads to a revised answer that is better grounded in the
    page evidence and local retrieval results.
    }
    \label{fig:qualitative_trace_case}
\end{figure*}

\subsection{Character Normalization}
\label{app:char_normalization}

The Char Normalizer converts raw OCR text into a more consistent form for
downstream reasoning while preserving the original OCR output. In our
implementation, this tool is based on the OpenCC Python library for
traditional-to-simplified Chinese conversion. The normalized text is used by
downstream extraction, retrieval, answer synthesis, and verification modules,
while the original OCR text remains available as page-level evidence.

\subsection{Term Extraction}
\label{app:term_extraction}

The Term Extractor identifies candidate terms from the question and page text.
It is implemented as a lightweight rule-based component rather than a separate
neural entity recognizer. The extractor collects quoted expressions,
term-like spans, and Chinese phrases from the question and page text, then
deduplicates them to form a candidate term list. These terms are used to build
retrieval queries for knowledge-intensive questions, especially Knowledge QA.

\subsection{Local Evidence Retriever}
\label{app:local_evidence}

The Local Evidence Retriever searches a frozen offline evidence pool,
constructed before evaluation for ancient-document QA. To build this pool, we
first collect candidate evidence through web search using terms derived from
the AncientDoc knowledge-QA setting, and then filter, clean, and consolidate
the returned content into short evidence snippets. Each entry is stored as a
key-value record, where the key corresponds to a term, entity, concept, book
title, or historical-cultural expression, and the value contains a concise
explanatory snippet with source metadata when available.

During inference, the evidence pool is fixed and never updated with test
examples or model outputs. This offline design serves two purposes. First, it
prevents uncontrolled external access during evaluation, making the evidence
path reproducible and inspectable. Second, it ensures that the main experiments
and ablation studies use the same external knowledge source, so performance
differences reflect the effect of the SAGE runtime rather than changes in
available evidence.

The retriever receives the question, extracted terms, and page context. It
always preserves the current page text as page-level evidence. For Knowledge
QA, it additionally retrieves up to eight local evidence snippets from the
offline pool. Retrieval uses transparent matching rules, including exact key
matching, substring matching, year-equivalent matching, key occurrence in the
question, and Chinese character overlap. Returned snippets are ranked,
deduplicated, and recorded with their evidence ids, matched queries, match
basis, scores, and source metadata.

\subsection{Budgets and Traceability}
\label{app:tool_budgets}

The runtime constrains tool use with task-specific budgets. The default maximum
number of tool calls is 3 for OCR, 5 for translation, 5 for reasoning QA, 7 for
Knowledge QA, and 6 for linguistic QA. The runtime rejects planned strategies
that exceed the budget or contain unregistered tools. Retrieval is also bounded:
each local evidence call returns at most eight KB snippets in addition to the
page context.

All tool invocations are recorded in the execution trace, including tool names,
inputs implied by the shared state, outputs, retrieved evidence, retrieval
statistics, and errors if any. These traces support the behavior analysis in the
main paper and allow us to diagnose how SAGE acquires evidence before answer
synthesis and verification.

\section{Qualitative Case Study}
\label{app:trace_case}

Figure~\ref{fig:qualitative_trace_case} presents a representative Knowledge QA
example from the execution traces. The input question asks about the
institutional background behind the practice of ``\textit{zhengwen kaoxian}''
(\textit{collecting texts and examining documents}). The Scholarly Planning
Agent first classifies the instance as Knowledge QA and plans an evidence path
that combines page reading, character normalization, term extraction, local
evidence retrieval, answer synthesis, and verification.

The Evidence-Grounded Execution Agent then reads the page using Qwen-VL-OCR,
normalizes the recognized text, extracts key terms, and retrieves local
evidence together with the page context. The initial candidate answer explains
the practice by referring to official historiography, civil-service selection,
and examination systems. However, the Scholar Verifier Agent finds that these
institutional claims are not directly supported by the retrieved evidence,
yielding an insufficient verdict with support rate $0.00$ and confidence
$0.50$.

Based on this feedback, SAGE triggers a replanning round. The revised answer
removes the unsupported institutional expansion and instead grounds the
explanation in the page evidence: the practice is framed as a scholarly activity
of collecting, organizing, and examining earlier documents to support moral
cultivation and textual judgment. After replanning, the verifier marks the
answer as supported, with support rate increasing from $0.00$ to $1.00$ and
confidence increasing from $0.50$ to $0.90$. This case illustrates how
\textbf{SAGE} uses verifier feedback to move from an over-extended candidate
answer toward a more evidence-grounded final response.

\end{document}